\documentclass[journal]{IEEEtran}
\usepackage{cite}
\ifCLASSINFOpdf
  \usepackage[pdftex]{graphicx}
\else
  \usepackage[dvips]{graphicx}
\fi
\usepackage{amsmath}
\usepackage{amsfonts}
\usepackage{tikz}
\usetikzlibrary{positioning}
\usepackage{booktabs}
\usepackage{multirow}
\usepackage{array}
\newcolumntype{M}[1]{>{\centering\arraybackslash}m{#1}}
\ifCLASSOPTIONcompsoc
 \usepackage[caption=false,font=normalsize,labelfont=sf,textfont=sf]{subfig}
\else
 \usepackage[caption=false,font=footnotesize]{subfig}
\fi

\usepackage{url}
\usepackage{xcolor,colortbl}
\definecolor{1best}{rgb}{1,1,.4}
\definecolor{2best}{rgb}{1,1,.6}
\definecolor{3best}{rgb}{1,1,.75}

\newcommand{\D}{\text{d}}
\newcommand{\T}{\mathsf{\scriptscriptstyle T}}
\newcommand{\defeq}{\,\raisebox{0.08ex}{:}\!\!=}
\newcommand{\E}{\mathrm{E}}
\newcommand{\Var}{\mathrm{Var}}

\begin{document}

\title{Color Constancy by Reweighting Image Feature Maps}
\author{Jueqin~Qiu,
        Haisong~Xu,
        and~Zhengnan~Ye% <-this % stops a space
\thanks{Jueqin Qiu, Haisong Xu, and Zhengnan Ye are with the State Key Laboratory of Modern Optical Instrumentation, College of Optical Science and Engineering, Zhejiang University, Hangzhou 310027, China. (e-mail: \{qiujueqin, chsxu\}@zju.edu.cn)}% <-this % stops a space
\thanks{Manuscript received xx, xxxx; revised xx, xxxx. (\textit{Corresponding author: Haisong Xu})}}

% The paper headers
\markboth{IEEE Transactions on Image Processing,~Vol.~, No.~xx, xx~xxxx}%
{Qiu \MakeLowercase{\textit{et al.}}: Color Constancy by Reweighting Image Feature Maps}
% The only time the second header will appear is for the odd numbered pages
% after the title page when using the twoside option.
% 
% *** Note that you probably will NOT want to include the author's ***
% *** name in the headers of peer review papers.                   ***
% You can use \ifCLASSOPTIONpeerreview for conditional compilation here if
% you desire.

\maketitle

% As a general rule, do not put math, special symbols or citations
% in the abstract or keywords.
\begin{abstract}
In this study, a novel illuminant color estimation framework is proposed for computational color constancy, which incorporates the high representational capacity of deep-learning-based models and the great interpretability of assumption-based models. The well-designed building block, feature map reweight unit (ReWU), helps to achieve comparative accuracy on benchmark datasets with respect to prior state-of-the-art deep learning based models while requiring more compact model size and cheaper computational cost. In addition to local color estimation, a confidence estimation branch is also included such that the model is able to simultaneously produce point estimate and its uncertainty estimate, which provides useful clues for local estimates aggregation and multiple illumination estimation. The source code and the dataset have been made available\footnote{https://github.com/QiuJueqin/Reweight-CC}.

\end{abstract}

% Note that keywords are not normally used for peerreview papers.

% For peerreview papers, this IEEEtran command inserts a page break and
% creates the second title. It will be ignored for other modes.
\IEEEpeerreviewmaketitle

\section{Introduction}
% The very first letter is a 2 line initial drop letter followed
% by the rest of the first word in caps.
% 
% form to use if the first word consists of a single letter:
% \IEEEPARstart{A}{demo} file is ....
% 
% form to use if you need the single drop letter followed by
% normal text (unknown if ever used by the IEEE):
% \IEEEPARstart{A}{}demo file is ....
% 
% Some journals put the first two words in caps:
% \IEEEPARstart{T}{his demo} file is ....
% 
% Here we have the typical use of a "T" for an initial drop letter
% and "HIS" in caps to complete the first word.
\IEEEPARstart{C}{olor} constancy of the human visual system is an essential prerequisite for many vision tasks, which compensates for the effect of the illumination on objects' color perception. Many computer vision applications are designed to extract comprehensive information from the \textit{intrinsic} colors of the objects, thereby requiring the input images to be color-unbiased. Unfortunately, the photosensors in modern digital cameras do not possess the ability of automatically compensating for the illuminant colors. To address this issue, a variety of computational color constancy algorithms have been proposed to mimic the dynamical adjustments of the cones in the human visual system~\cite{Brainard:1986, Lam:2005, Zhang:2016}. 

Computational color constancy generally works by first estimating the illuminant color, and then compensating it by multiplying the reciprocal of the illuminant color to the color-biased image. Existing computational color constancy algorithms can be classified into \textit{a priori} assumption-based ones and learning-based ones, according to whether a training process is needed. Typical assumption-based algorithms include Gray-World~\cite{Buchsbaum:1980}, White-Patch~\cite{Brainard:1986}, variants of Gray-Edge~\cite{Weijer:2005, Weijer:2007a}, and some that utilize statistical information of the images~\cite{Hannah:2004}. Although assumption-based methods are lightweight and comprehensible, their performances are likely to decrease dramatically if these restrictive assumptions are not satisfied. Learning-based algorithms can be further grouped into low-level ones and high-level ones. Typical low-level methods include Color-by-Correlation~\cite{Finlayson:2001}, Gamut Mapping~\cite{Barnard:2000}, Bayesian color constancy~\cite{Brainard:1997}, etc. Since the spatial and textural information has been lost when generating low-level color descriptors, these methods are prone to produce ambiguous estimates if they have not ``seen'' the colorimetric patterns of the test images in the training phase. In recent years, following the massive success of deep learning in computer vision community, high-level color constancy algorithms based on the convolutional neural network (CNN) have achieved state-of-the-art performances on the benchmark datasets~\cite{Cheng:2015, Bianco:2015, Hu:2017a}. However, we notice that many existing CNN-based models adopt the architectures from image classification domain, e.g., AlexNet~\cite{Krizhevsky:2012} in \cite{Mahmoud:2018}, SqueezeNet~\cite{Iandola:2016} in \cite{Hu:2017a}, which are overcomplicated and inefficient when dealing with color-relevant applications.

Compared to visual recognition or understanding tasks, color constancy has its own unique properties:

\noindent\textbf{High priority} - It is generally suggested that the ganglion cells within the retina are the basis of color processing mechanisms of human color constancy, which respond to the activations of cone photoreceptors at the very first stage of the human visual system~\cite{Zhang:2016, Kolb:2003}. Therefore, it makes more sense to perform the computational color constancy before the explorations of further semantic information.

\noindent\textbf{Condensation and superficiality} - The useful cues for color constancy are highly condensed: if there exists neutral or specular regions in the image, computational color constancy can be done by pooling these regions in the spatial domain~\cite{Shi:2012, Joze:2012, Cheng:2014}; if there exists shading regions, it can be done by extracting these regions in the gradient domain~\cite{Lei:2016}. Instead of learning very deep features as many visual recognition tasks try to do, color constancy is such a task that can be solved in a more superficial level.

Based on these observations, in this study an efficacious building block, dubbed \textit{feature map reweight unit (ReWU)}, is proposed to extract informative cues from image feature maps. With ReWUs, the illuminant colors can be accurately estimated with only 1--3 convolutional layers, which makes our model more compact and efficient compared to prior CNN-based models.

\section{Preliminaries}

Assuming Lambertian reflection and uniform illumination, the raw intensity vector $\boldsymbol{p}(x)\in\mathbb{R}^3$ at pixel $x$ recorded by a typical tri-chromatic photosensor is given by~\cite{Horn:1979}
\begin{equation}
\label{eq:response}
\boldsymbol{p}(x) = \kappa(X)\int\limits_\Omega\!\boldsymbol{\rho}(\lambda,X)\,\boldsymbol{\ell}(\lambda)\,\boldsymbol{s}(\lambda)\,\D{\lambda}\,,
\end{equation}
where $X$ is the point in the space corresponding to $x$, $\boldsymbol{\rho}(\lambda,X)$ is the spectral reflectance at $X$, $\boldsymbol{\ell}(\lambda)$ is the spectral irradiance by an arbitrary incident illumination, $\boldsymbol{s}(\lambda)$ denotes the spectral sensitivities of the photosensors, and $\kappa(X)$ is a geometry-dependent factor.

Recovering $\boldsymbol{\ell}$ from the integration is an ill-posed problem. A common approximation is to reformulate \eqref{eq:response} with a Hadamard product as per von Kries model~\cite{Brainard:1986}:
\begin{equation}
\label{eq:responsesimple}
\boldsymbol{p}(x)\approx\boldsymbol{p}^\ast(x)\circ\mathbf{L}\,,
\end{equation}
where $\boldsymbol{p}^\ast(x)$ is the vector of the ``intrinsic'' color of the object when it is observed under a canonical illuminant, and $\mathbf{L}\in\mathbb{R}^3$ is the color vector of the arbitrary light source: $\mathbf{L} = \int\nolimits_\Omega\!\boldsymbol{\ell}(\lambda)\,\boldsymbol{s}(\lambda)\,\D{\lambda}\,$.

In this way, computational color constancy is equivalent to illuminant color estimation problem and consequently can be achieved by compensating for the color of the arbitrary light source:
\begin{equation}
\label{eq:groundtruthresponse}
\boldsymbol{p}^\ast(x)\approx\boldsymbol{p}(x)\circ\mathbf{L}^{-1}\,,
\end{equation}
where $(\cdot)^{-1}$ is the element-wise reciprocal calculation.

\section{Model}\label{sec:methodology_and_models}

In this section we first introduce the \textit{feature map reweight unit (ReWU)}, a lightweight build block specifically designed for color-relevant vision tasks. ReWUs work cooperatively with hierarchical feature maps extracted by the convolution blocks. Taking one feature map as input, a ReWU extracts informative cues for illuminant color estimation by assigning varying weights to the pixels in the feature map. Pixels with higher weights will dominate the decision of color estimation in the downstream of the network, and those with lower weights will be suppressed or even completely ignored. Since generating feature maps can be regarded as weighting the input image with different kernels, we denominate the proposed technique \textit{pixel reweighting}.

In subsection B, the network architecture for illuminant color estimation is proposed, which can be interpreted as connecting up a set of ReWUs to hierarchical convolution blocks in a sequential convolutional neural network. To yield final illuminant color estimate, the global average poolings~\cite{Lin:2013} are applied to each reweighted feature map such that only the pixels with highest activations are kept and fed into final fully-connected layers for color regression. The proposed model takes color-biased \textit{image patches} as inputs, thus it works for both local and global estimation tasks.

In addition to the accuracy metric, it is also suggested that the uncertainty information of illuminant estimation is of great importance for practical uses~\cite{Barron:2017}. To circumvent the absence of uncertainties prediction in the naive model, in subsection C, we add a \textit{confidence estimation branch} to the network such that the point estimates (illuminant colors) as well as the uncertainty estimates can be produced simultaneously.

\subsection{Feature map reweight unit (ReWU)}

Based on the observation that some pixels matter more than others in inferring illuminant color~\cite{Joze:2012}, many algorithms were designed to detect and analyze the pixels that satisfy some constraints~\cite{Liu:1995, Weng:2005}. In the very simplest case, in order to extract all \textit{near-achromatic} pixels that might belong to the neutral objects, the selective Gray Point algorithm~\cite{Huo:2006} loops through all pixels $x$'s in an input image and finds those located within a specified region centering at the neutral point:
\begin{equation}
\label{eq:pixelconstraint}
x\defeq%
\begin{cases}
achromatic & \text{if }\left|(u_x-u_0)-(v_x-v_0)\right|\leq T\\
& \text{and} \left|(u_x-u_0)+(v_x-v_0)\right|\leq T\,,\\
chromatic & \text{otherwise}\,,
\end{cases}
\end{equation}
where $[u_x, v_x]=[r_x/(r_x+g_x+b_x), g_x/(r_x+g_x+b_x)]$ is the dimension-reduced chromaticity coordinate of pixel $x$, $[u_0, v_0]$ is the coordinate of the neutral point, which is predifined given a particular camera and the reference illuminant, and $T$ is an adjustable distance threshold. 

Equation~\eqref{eq:pixelconstraint} can be rewritten in a more computational friendly form such that it can be easily handled by the tensor manipulations in the deep-learning frameworks:
\begin{equation}
\label{eq:pixelconstraint_matrix0}
x\defeq%
\begin{cases}
achromatic & \text{if }\mathcal{W}(u_x, v_x)>0\,,\\
chromatic & \text{otherwise}\,,
\end{cases}\\[.8em]
\end{equation}
where

\begin{equation}
\label{eq:pixelconstraint_matrix}
\mathcal{W}(u_x, v_x) = ReLU\left(min\left(\mathbf{A}[u_x, v_x]^\T+\mathbf{b}\right)\right)\,,\\[.8em]
\end{equation}
and

\begin{equation}
\label{eq:pixelconstraint_matrix_ab}
\mathbf{A}=\begin{bmatrix}%
1 & -1\\
-1 & 1\\
1 & 1\\
-1 & -1
\end{bmatrix},\quad\mathbf{b} = \begin{bmatrix}%
T-u_0+v_0\\
T+u_0-v_0\\
T-u_0-v_0\\
T+u_0+v_0
\end{bmatrix}\,.
\end{equation}

\begin{figure}[!t]
\centering
\includegraphics[width=\linewidth]{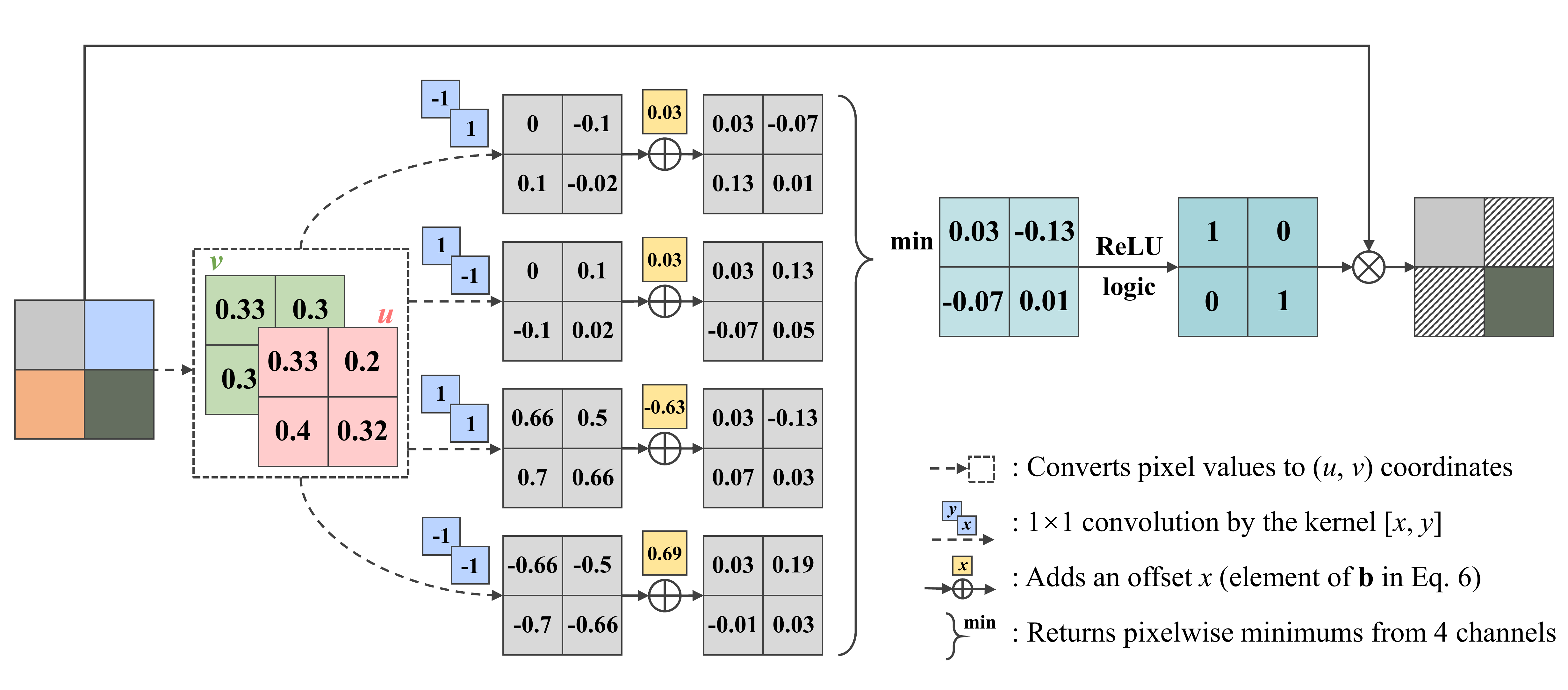}
\caption{Visualization of near-achromatic pixels extraction by the function $\mathcal{W}$ in Eq.~\eqref{eq:pixelconstraint_matrix}. Pixels with chromaticity coordinates $(u, v)$ located within a small region (controlled by the threshold $T$) centering at the neutral point $(u_0, v_0)$ will be considered as near-achromatic. For the sake of simplicity, we use $(u_0, v_0)=(0.33, 0.33)$ and $T=0.03$ in this example.}
\label{fig:achromaticpixels}
\end{figure}

In Eq.~\eqref{eq:pixelconstraint_matrix} $min(\cdot)$ operator returns the minimum element in the input vector, and $ReLU(\cdot)$ is the rectified linear units~\cite{Nair:2010} that clamps all the negative values at zero. Figure~\ref{fig:achromaticpixels} demonstrates how the function $\mathcal{W}$ works in this example to extract two near-achromatic pixels from a quad.

Inspired by this selective extraction mechanism, we propose the \textit{feature map reweight unit (ReWU)} that is able to highlight informative pixels and suppress less useful ones by supervisely learning how to impose ``constraints'' and ``thresholds'' on the feature maps. Instead of using the simple yes-or-no scheme in Eq.~\eqref{eq:pixelconstraint} to select pixels, ReWUs assign varying weights to pixels in the input feature maps.

\begin{figure}[!t]
\centering
\includegraphics[width=\linewidth]{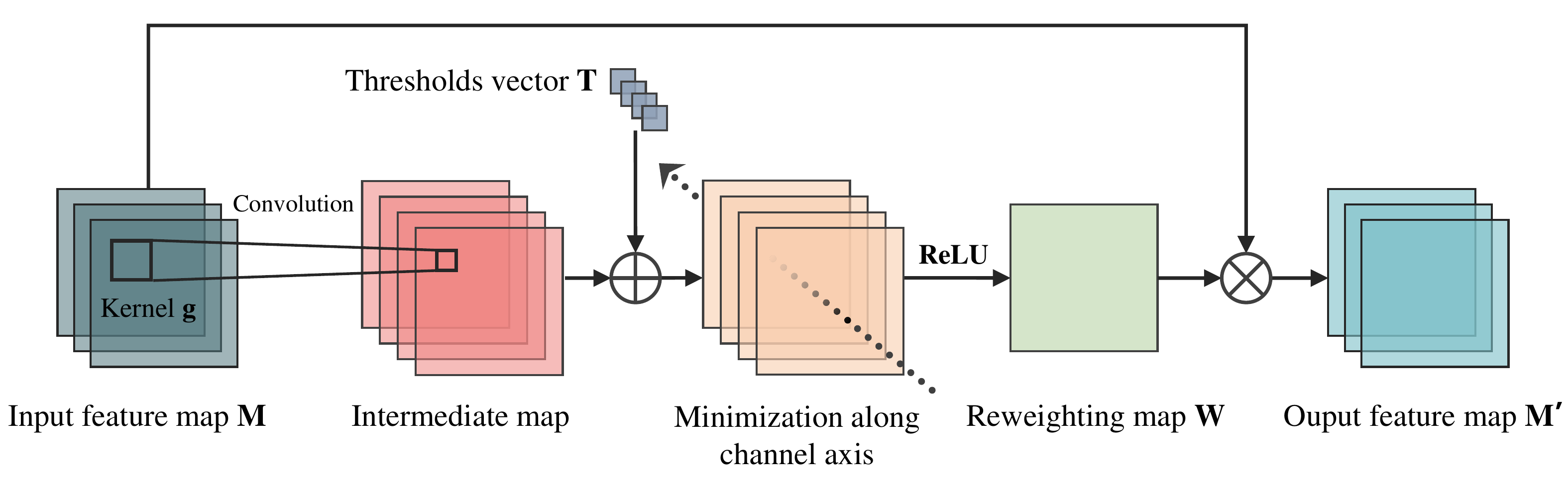}
\caption{Schematic structure of the feature map reweight unit (ReWU). The trainable scaling factor $\alpha$ is omitted.}
\label{fig:microarchitecture}
\end{figure}

Formally, the reweight unit $\mathcal{W}: \mathbf{M}\in\mathbb{R}^{H\times{}W\times{}C}\rightarrow\mathbf{M}^\prime\in\mathbb{R}^{H\times W\times C}$ produces the reweighted output feature map $\mathbf{M^\prime}$ by pixelwise multiplying the input feature map $\mathbf{M}$ with a \textit{reweighting map} $\mathbf{\tilde{W}}$:
\begin{equation}
\label{eq:outputfeaturemap}
\mathbf{M^\prime} = \mathbf{\tilde{W}}\circ\mathbf{M}\,,
\end{equation}
where $\mathbf{\tilde{W}}$ is the dimension-expanded version of $\mathbf{W}$ that contains $K$ copies of $\mathbf{W}$ along channel axis, and $\mathbf{W}$ is calculated by first convolving the input feature map with $1\times1$ kernels and activating those pixels that satisfy the constraints implicitly embbed in the kernels $\mathbf{g}$ and thresholds $\mathbf{T}$:
\begin{equation}
\label{eq:reweightmap}
\mathbf{W} = \alpha\cdot{}ReLU\left(min\left(\mathbf{g}\ast\mathbf{M}+\mathbf{T}\right)\right)\,.
\end{equation}
Here $\mathbf{g}$ is a $1\times1\times{}C\times{}K$ kernel tensor, where $C$ is the number of channels in the input feature map, and $K$ is the number of kernels, which can also be interpreted as \textit{the number of constraints} imposed on the feature map. $\mathbf{T}$ is a $K\times1$ thresholds vector that determines the tolerance for the pixels being away from the ``hotspot'' in the $C$-dimensional space. Symbol $\ast$ denotes the spatial convolution, $min(\cdot)$ herein is the minimization operation \textit{along the channel axis}, and $\alpha$ is a trainable scaling factor that allows to adjust the intensity of the reweighting map. The schematic structure of ReWU is illustrated in \figurename~\ref{fig:microarchitecture}.

Apparently, the example of near-achromatic pixels extraction in \figurename~\ref{fig:achromaticpixels} is a particular instance of ReWU where 4 constraints are imposed upon the 2-dimensional chromaticity plane, produceing a binarized reweighting map.

The ReWUs are highly parameter-economic. In our experiments, by setting the number of kernels to be equal to the number of channels in the input feature maps ($K=C$), a ReWU has only $O(C^2)$ space complexity, which reduces the numbers of parameters by 1--2 order of magnitude compared to a conventional convolution blocks.

It should be noted that although the $1\times1$ convolution has been used in other network structures like Inception~\cite{Szegedy:2015}, ShuffleNet~\cite{Zhang:2017} and LiteFlowNet~\cite{Hui:2018}, the idea behind the ReWU is quite different. Furthermore, it is also possible to modify $1\times1$ kernels in ReWUs to others with larger receptive fields, if the structural information among neighboring pixels are expected to be utilized.

\subsection{Network architecture}\label{sec:architecture}

\begin{figure}[!t]
\centering
\includegraphics[width=\linewidth]{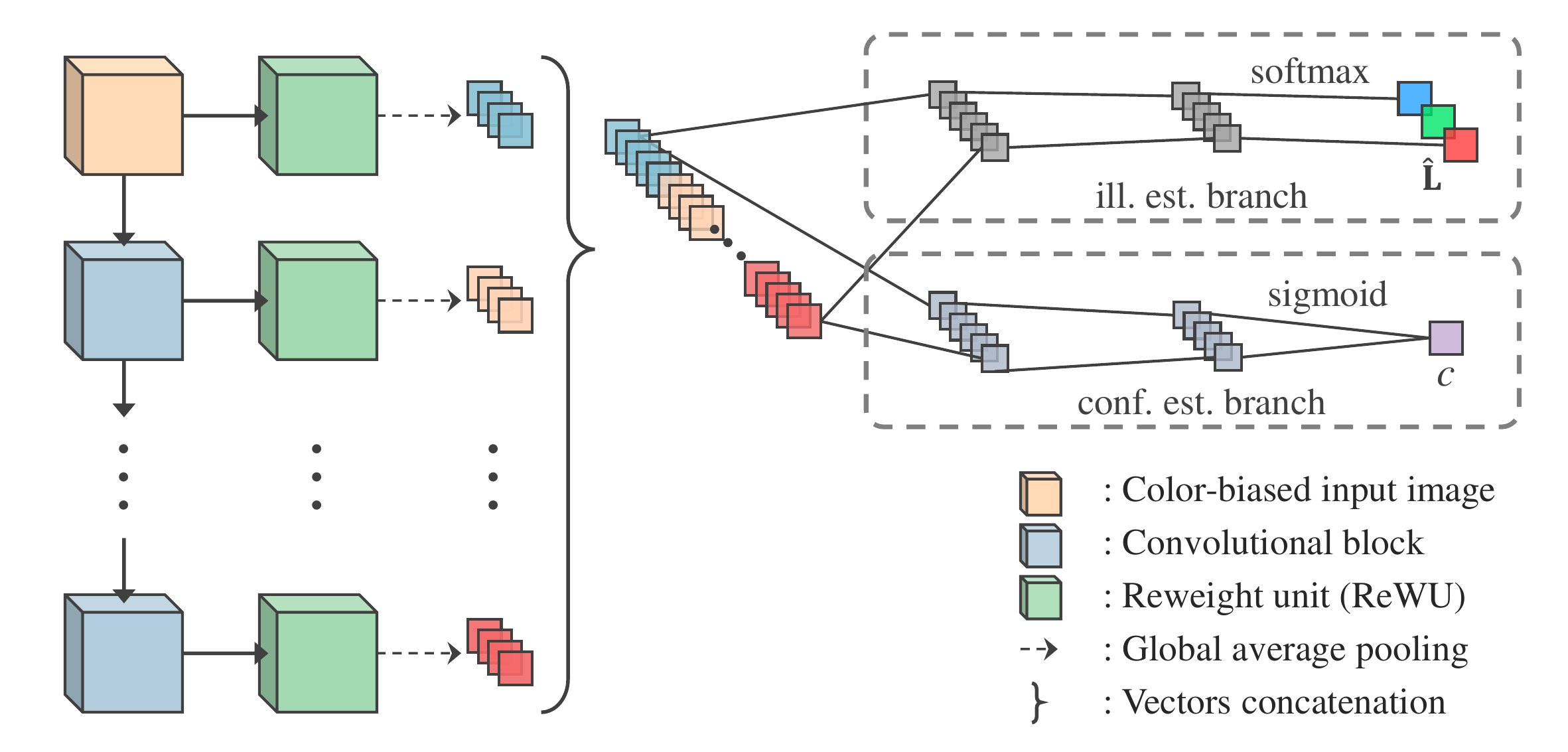}
\caption{The architecture of the proposed illuminant color regression network. The \textit{illuminant estimation branch} produces an RGB triplet that represents the estimated illuminant color, and the \textit{confidence estimation branch} produces a single score that represents the confidence of the network about its color estimate.}
\label{fig:architecture}
\end{figure}

The proposed illuminant color estimation network is a CNN-based regression model that takes the color-biased image patches as inputs. The network is built up by connecting up a set of ReWUs to the hierarchical feature maps (including the color-biased input image itself) in a sequential convolutional neural network. For each ReWU, the global average pooling~\cite{Lin:2013} is applied to its output feature map $\mathbf{M^\prime}$ such that only pixels with highest activations in respective channels are kept and collected into an \textit{activation vector}. Activation vectors from different ReWUs are concatenated into a long vector, which is fed into the fully-connected illuminant color estimation branch to produce final RGB triplet regressors $\hat{\mathbf{L}}$. Figure~\ref{fig:architecture} illustrates the abstract architecture of the proposed regression network. The network can be trained end-to-end by using the mean squared errors between the estimated and the ground truth illuminant colors (both $\ell_1$-normalized) as the loss function\footnote{We tried to optimize the network towards minimizing both MSE and cosine error but found no noticeable difference in performance. However, minimizing MSE accelerated the convergence approximately 20\% over the cosine error.}. To produce the $\ell_1$-normalized estimates, we add an extra softmax layer after the final fully-connected layer.

The number of hierarchical levels of the network can be flexibly chosen. We found that stacking only 1--3 convolution blocks is adequate to achieve the practical accuracy for illuminant color estimation. More ablation experiments about the depths of the models will be given in section~\ref{sec:experiments}. 

Cooperating the global average poolings with ReWUs makes our model interpretable. Applying global average poolings to the output feature maps of the ReWUs is equivalent to applying \textit{weighted poolings} to the convolutional feature maps or the input image. For example, assuming that a ReWU highlights the human face regions in the input image, the global average pooling acts like the method in \cite{Bianco:2012} as it assigns high weights to the pixels with skin colors. Looked at from another point, the ReWUs and global average poolings act somewhat like the Squeeze-and-Excitation blocks~\cite{Hu:2017a} or attention modules~\cite{Wang:2017}, but in a more extreme way that only the most important cues could participate in the decision-making of the illuminant color estimation.

It is also worth noting that the global average pooling makes our network available to be adapted to arbitrary input sizes, as the length of the concatenated vector depends only on the sum of channel amounts from all reweighted feature maps, which can be determined in advance. Furthermore, both ReWU and global average pooling are shift invariant, which is reasonably required for the illuminant color estimation task.

\subsection{Confidence estimation branch}

In making decisions, it is often necessary to be able to tell whether a model is certain about its outputs. To this end, we add an extra \textit{confidence estimation branch} to the naive network dicussed above, such that the predictive uncertainties can be produced simultaneously along with the point estimates~\cite{Pearce:2018}. 

The proposed confidence estimation branch is motivated by the out-of-distribution detecting algorithm in \cite{DeVries:2018}. We adapt it to make it fit for regression tasks. The branch will mainly focus on the aleatoric uncertainties that come from the out-of-distribution data or foreign situations, but neglect the epistemic uncertainty that comes from the intrinsic nature of the model.

The confidence estimation branch is added to the naive network \textit{in parallel} with the fully-connected illuminant color estimation branch, as depicted in \figurename~\ref{fig:architecture}, such that both braches receive the same input (the concatenated vector). The confidence estimation branch outputs a single scalar $c\in[0, 1]$ parametrized by the sigmoid function, which represents the network's confidence that the illuminant color can be correctly estimated given a color-biased input image. Ideally, if the network is confident about its illuminant color estimate, $c$ should be close to 1. Conversely, if the network is not confident that it can produce an accurate estimate, $c$ should be close to 0.

We use the adjusted mean squared error as the \textit{task loss} in this new network:
\begin{equation}
\label{eq:taskloss}
\mathcal{L}_t = \lVert{\mathbf{L}^\ast-\hat{\mathbf{L}}^\prime}\rVert_2^2\,,
\end{equation}
where $\mathbf{L}^\ast$ is the ground truth color vector of the light source (also $\ell_1$ normalized), and $\hat{\mathbf{L}}^\prime$ is the adjusted estimated color vector produced by interpolating between the original estimate $\hat{\mathbf{L}}$ and the ground truth, where the degree of interpolation is indicated by the confidence:
\begin{equation}
\label{eq:adjustedcolorvector}
\hat{\mathbf{L}}^\prime = c\cdot\hat{\mathbf{L}} + (1-c)\cdot\mathbf{L}^\ast\,.
\end{equation}

Given an input image for which the illuminant color estimation branch might produce large estimation error, assigning it with a lower confidence will give the network more ``hints'' to reduce the task loss. To prevent the network from minimizing the task loss by always choosing $c = 0$ and receiving the entire ground truth, we add the \textit{regularization loss} as the penalty:
\begin{equation}
\label{eq:regularizationloss}
\mathcal{L}_r = -\log(c).
\end{equation}

The final loss function is the weighted sum of the task loss and the regularization loss:
\begin{equation}
\label{eq:finalloss}
\mathcal{L} = \mathcal{L}_t + \lambda\mathcal{L}_r\,.
\end{equation}
where $\lambda$ is a hyperparameter to balance the magnitudes of the two terms.

Giving a $c$ smaller than 1 will push the orginal vector $\hat{\mathbf{L}}$ closer to the target one, resulting in a reduction in the task loss at the cost of an increase in the regularization loss. Optimizing the overall network is like running a competition game, wherein the network can reduce its final loss only if it can successfully predict which inputs it is likely to be accurately estimated, and assigns hight confidence scores to them.

Under the uniform illumination assumption, $c$'s from different input image patches can be treated as weights to aggregate local illuminant color estimates into a global estimate. When multiple illuminants exist in the image, the uncertainty estimates help to ameliorate the perturbation of color estimates across different local patches. In addition, quantifying predictive uncertainty in the networks also allows for better informed decisions. For example, given a corner case, if the confidences from all local estimates are lower than a threshold, it is considerable to run some fallback algorithms to get more conservative result.

\section{Experiments}\label{sec:experiments}

\subsection{Incarnate networks and hyperparameters}

Based on the basic architecture proposed in section~\ref{sec:architecture}, three incarnate networks were built up and tested, with 1, 2, and 3 hierarchical levels respectively. Given a specified number of levels, two variants, one with and another without the confidence estimation branch, were also compared.

For the conventional convolution blocks in the networks, we adopted the designs in Inception v3 model~\cite{Szegedy:2016} and set the numbers of kernels to 32, 32, and 64 for the convolutional layers in 3 hierarchical levels respectively. For each ReWU, we fixed the number of $1\times1$ kernels to be equal to the number of channels in its input feature maps, i.e., $K=C$. One exception was the first ReWU directly connected to the color-biased input image, which had 16 kernels. Table~\ref{tab:incarnatenetworks} lists the detailed architectures for three naive networks without the confidence estimation branches.

\begin{table*}[!t]
\setlength\aboverulesep{0pt}
\setlength\belowrulesep{0pt}
\setlength{\tabcolsep}{0.36em}
\renewcommand{\arraystretch}{1.3}
\caption{Architecture details of incarnate networks with 1, 2, and 3 hierarchical levels. The confidence estimation branches are not included but they share the same architectures as the illuminant estimation branches, only with the final 3-neuron layer replaced by a 1-neuron one.}
\label{tab:incarnatenetworks}
\centering
\begin{tabular}{M{1.4cm}||M{1.35cm}|M{1.35cm}|M{1.35cm}||M{1.4cm}|M{1.4cm}|M{1.4cm}||M{1.5cm}|M{1.5cm}|M{1.5cm}}
\toprule
& \multicolumn{3}{c||}{1-Hierarchy} & \multicolumn{3}{c||}{2-Hierarchy} & \multicolumn{3}{c}{3-Hierarchy}\\
\hline
layer name & output size & Conv\,/\,FC & ReWU & output size & Conv\,/\,FC & ReWU & output size & Conv\,/\,FC & ReWU\\
\hline
input (hrchy\_0) & $224\times224$ & -- & $1\times1$, 16 & $224\times224$ & -- & $1\times1$, 16 & $224\times224$ & -- & $1\times1$, 16\\
\cline{1-10} 
hrchy\_1 & $112\times112$ & $3\times3$, 32 stride 2 & $1\times1$, 32 & $112\times112$ & $3\times3$, 32 stride 2 & $1\times1$, 32 & $112\times112$ & $3\times3$, 32 stride 2 & $1\times1$, 32\\
\cline{1-10} 
hrchy\_2 & \multirow{2}{*}{--} & \multirow{2}{*}{--} & \multirow{2}{*}{--} & $112\times112$ & $3\times3$, 32 & $1\times1$, 32 & $112\times112$ & $3\times3$, 32 & $1\times1$, 32\\
\cline{1-1} \cline{5-10} 
hrchy\_3 & & & & -- & -- & -- & $112\times112$ & $3\times3$, 64 & $1\times1$, 64\\
\cline{1-10} 
concat & $35\times1$ & \multicolumn{2}{c||}{--} & $67\times1$ & \multicolumn{2}{c||}{--} & $131\times1$ & \multicolumn{2}{c}{--}\\
\cline{1-10}
fc\_1 & $64\times1$ & 1, 64 & \multirow{4}{*}{--} & $128\times1$ & 1, 128 & \multirow{4}{*}{--} & $256\times1$ & 1, 256 & \multirow{4}{*}{--}\\
\cline{1-3} \cline{5-6} \cline{8-9}
fc\_2 & $32\times1$ & 1, 32 & & $64\times1$ & 1, 64 & & $128\times1$ & 1, 128 &\\
\cline{1-3} \cline{5-6} \cline{8-9}
fc\_3 & $16\times1$ & 1, 16 & & $32\times1$ & 1, 32 & & $64\times1$ & 1, 64 &\\
\cline{1-3} \cline{5-6} \cline{8-9}
estimate & $3\times1$ & 1, 3 & & $3\times1$ & 1, 3 & & $3\times1$ & 1, 3 &\\
\bottomrule
\end{tabular}
\end{table*}

\subsection{Initialization}\label{sec:initialization}

For an incarnate network, the kernels in the conventional convolution blocks were initialized with the parameters from the corresponding layers in Inception v3 model pre-trained on ImageNet classification task~\cite{Szegedy:2016}. For the fully-connected layers in both branches, the random normal initializers were used.

The initialization for the kernels in ReWUs is a bit more tricky. Let us first consider such an example that the pixel values in the input feature map $\mathbf{M}$ have uniform distribution between 0 and 1 and that the values of different channels in a certain pixel are independent and identically distributed. If the kernels in the tensor $\mathbf{g}$ are randomly initialized from a standard gaussian and the thresholds vector $\mathbf{T}$ is zero-initialized, we have
\begin{equation}
\label{eq:expectation}
\E[Z] = \E[G]\,\E[M] = 0\,,\\
\end{equation}
\begin{equation}
\label{eq:variance}
\begin{aligned}
\Var[Z]= C\Big(&\Var[G]\,\Var[M] + \Var[G]\,\E^2[M] +\Big. \\
\Big.&\Var[M]\,\E^2[G]\Big)=\frac{C}{3}\,,
\end{aligned}
\end{equation}
where $G$, $M$, and $Z$ represent the random variables of values in $\mathbf{g}$, $\mathbf{M}$, and $(\mathbf{g}\ast\mathbf{M})$ respectively, $C$ is the number of channels in $\mathbf{M}$, and $\E[\cdot]$ and $\Var[\cdot]$ denote the expectation and variance calculations. Consequently, the random variable in $(\mathbf{g}\ast\mathbf{M})$ is also normally distributed with $\mu=0$ and $\sigma=\sqrt{C/3}$, which will lead to a zeros-tensor output after ReLU, because minimizing $(\mathbf{g}\ast\mathbf{M})$ along the channel axis will always yield negative values.

We address this issue by appropriately initializing $\mathbf{T}$ with a constant $\tau$ (i.e., $\mathbf{T}=\tau\cdot ones(H,W,C)$ where $ones(\cdots\!\,)$ is a tensor with specified size filled with 1's) so as to ensure \textit{approximate 50\% of elements in the initial reweighting map $\mathbf{W}$ are non-zero}, given input feature maps with arbitrary dimensionalities and distributions. 

First we apply a normalization function $C\!N(\cdot)$, dubbed channel normalization, to $(\mathbf{g}\ast\mathbf{M})$ such that the random variable in $C\!N(\mathbf{g}\ast\mathbf{M})$ has zero mean and unit standard deviation \textit{along the channel axis}. Channel normalization performs computation to values in the intermediate map $(\mathbf{g}\ast\mathbf{M})$:
\begin{equation}
\label{eq:channelnormalization}
\hat{p}_{x,i} = \frac{p_{x,i}-\mu_x}{\sigma_x+\epsilon}\,,
\end{equation}
where $x$ is the 2D spatial index, $i\in\{1,\ldots,K\}$ is the channel index, and $\epsilon$ is a small constant to avoid instability. $\mu_x$ and $\sigma_x$ are the mean and standard deviation \textit{individually calculated for each pixel} over its all channels:
\begin{equation}
\label{eq:meanandstd}
\mu_x = \frac{1}{K}\sum_{i=1}^{K}p_{x,i}\,,\quad\sigma_x=\sqrt{\frac{1}{K}\sum_{i=1}^{K}(p_{x,i}-\mu_x)^2}\,.
\end{equation}

\begin{figure}[!t]
\includegraphics[width=\linewidth]{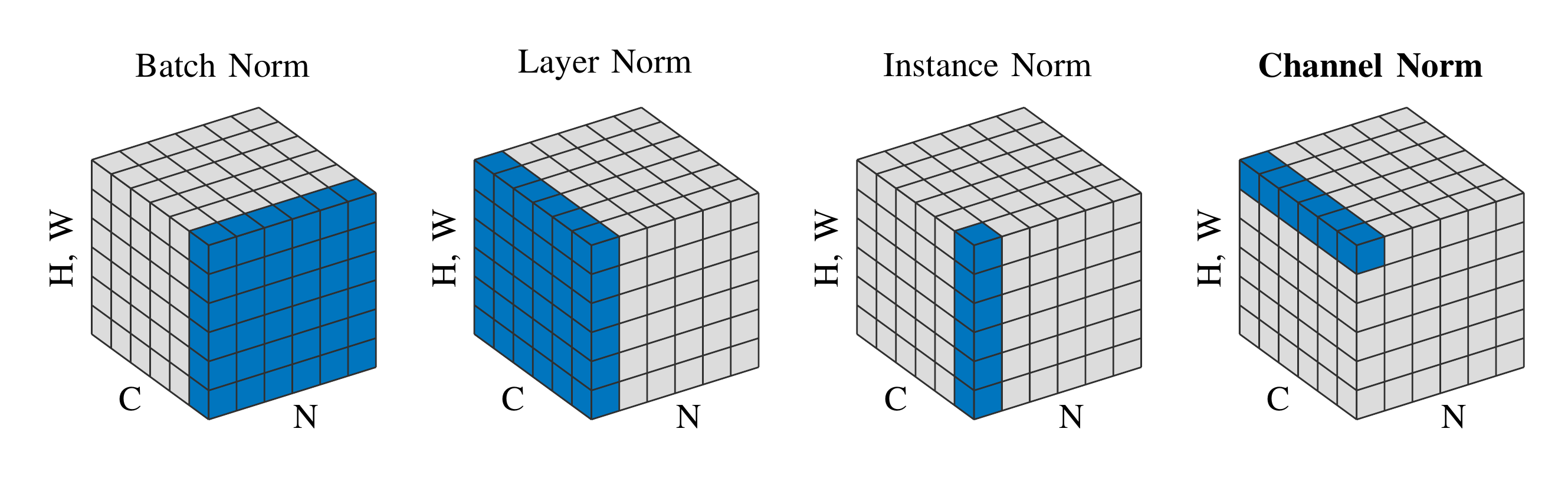}
\caption{Comparison of channel normalization with batch normalization~\cite{Ioffe:2015}, layer normalization~\cite{Ba:2016}, and instance normalization~\cite{Ulyanov:2016}. Each subplot shows a feature map tensor, with $N$ as the batch axis, $C$ as the channel axis, and $(H, W)$ as the spatial axes. The pixels in blue are normalized by the same mean and standard deviation. The figure is motivated by \cite{Wu:2018}.}
\label{fig:channelnormalization}
\end{figure}

Figure~\ref{fig:channelnormalization} compares the channel normalization with other existing feature normalization methods.

To ensure half of elements in $\mathbf{W}$ are non-zero, the intermediate 1-channel map $min(C\!N(\mathbf{g}\ast\mathbf{M})+\mathbf{T})$ should have an expectation of zero because its elements have approximately symmetrical distribution:

\begin{equation}
\label{eq:init1}
\begin{aligned}
&\E[min(C\!N(\mathbf{g}\ast\mathbf{M}))+\mathbf{T}]\\
=&\E[min(C\!N(\mathbf{g}\ast\mathbf{M}))+\tau]=0\,.
\end{aligned}
\end{equation}
Therefore,
\begin{equation}
\label{eq:init2}
\begin{aligned}
\tau&= \E[\tau] = -\E[min(C\!N(\mathbf{g}\ast\mathbf{M}))]\\
 &= -K\int_{-\infty}^\infty{}z\,\varphi(z;0,1)\big(1-\Phi(z;0,1)\big)^{(K-1)}\D{z}\,,
\end{aligned}
\end{equation}
where $\varphi(\cdot;0,1)$ and $\Phi(\cdot;0,1)$ are the cumulative distribution and probability density functions of the standard normal distribution. The proof of \eqref{eq:init2} is presented in the Appendix.

\subsection{Benchmark datasets and images pre-processing}\label{sec:benchmark_datasets}

We performed intra- and inter-camera experiments to evaluate the performances of the proposed method. For intra-camera evaluation, ColorChecker RECommended dataset~\cite{Hemrit:2018} and NUS-8 Camera dataset~\cite{Cheng:2014} were used for benchmarking, in which 10 camera models in total were individually tested by three-fold cross validation. ColorChecker RECommended dataset is an updated version of the original Gehler-Shi dataset~\cite{Gehler:2016}, in which a new ``recommended'' ground-truth set was re-generated. These updated ground-truths were used in our experiments as the labels to evaluate the illuminant estimation accuracy for Canon 1D and Canon 5D camera models. Following previous works~\cite{Bianco:2015, Hu:2017a, Bianco:2017}, in intra-camera evaluations, we used the device-dependent raw images as the inputs for both training and validation, after black levels subtracted.

For inter-camera accuracy evaluation, in addition to the 10 camera models from the aforementioned two standard datasets, we also included Cube dataset~\cite{Banic:2017} for benchmarking, which contains 1365 exclusively outdoor images taken by a Canon 550D camera. We named this merged dataset as \textit{MultiCam dataset}. The validation of the generalization abilities of our models was carried out using the leave-one-out cross validation on the MultiCam dataset. Specifically speaking, in each training-validation experiment, we used the images from 10 out of 11 camera models as the training data and computed statistics on the images from the rest one. By repeating this procedure 11 times and calculating the arithmetic mean, final inter-camera illuminant estimation accuracy was obtained.

Since different photosensors record quite distinct responses given the same incident stimulus, it is necessary to pre-process the raw images from different camera models such that all the training images are in a common ``calibrated'' color space. We accomplished this by converting raw images (black levels subtracted) from device-dependent color spaces into the standard linear sRGB color space using $3\times3$ color correction matrices~\cite{Coffin:2017}. Since the channel sensitivity scaling coefficients have already been implicitly included in the color correction matrices, the ground truths of illuminant colors were also adjusted accordingly to fit the calibrated images. The conversion to linear sRGB color space does not preserve the distances between illuminants, therefore to make the results comparable, the estimated illuminant colors were converted back into the individual device-dependent color spaces and then the angular errors with respect to the \textit{unadjusted} ground truths were calculated.

To address the problem that the amount of training data were too small, especially for intra-camera experiments, to train overfitting-free models, data augmentation was performed by randomly cropping square sub-images from the original full-resolution images and rotating it with a random degree. Depending on the orginal image sizes, the crops had sizes from $256\times256$ to $768\times768$. Before being fed into a network, all sub-images would be resized to $224\times224$. Besides, to make our networks more robust to unseen scenes, given each sub-image, we intentionally added a small random bias with 50\% probability to the illuminant color. The bias was restricted not to be greater than $5^\circ$ in angular error, and not to deviate from the ground-truth over 0.006 unit of $D_{uv}$~\footnote{$D_{uv}$ was calculated after converting device-dependent illuminant RGB into the CIE 1960 UCS~\cite{Fryc:2005}. We imposed this constraint to guarantee that the generated illuminant would locate at a point being not too far away from the orginal one in the direction of the iso-temperature line, which was close to the practical situations. 0.006 is an empirical threshold that determines how far a ``white light'' is allowed to deviate from the Planckian locus~\cite{ANSI:2011}.}.

\subsection{Implementation details}

The proposed networks were implemented in TensorFlow~\cite{TensorFlow:2015}. Nadam~\cite{Dozat:2016} was employed as the optimizer with a base learning rate of $5\times10^{-5}$. When plateaus detected, the learning rate would be decreased by 10\%. Dropout~\cite{Srivastava:2014} with probability of 0.2 was included for the fully-connected layers.

For a naive network (i.e., a network without the confidence estimation branch), all parameters in ReWUs, convolutional blocks and fully-connected layers are free to vary during training; for a network with the confidence estimation branch, we trained the naive network first, then freezed all the parameters and trained the confidence branch separately. The hyperparameter $\lambda$ in \eqref{eq:finalloss} was initialized such that the task loss $\mathcal{L}_t$ and the balanced regularization loss $\lambda\mathcal{L}_r$ were of the same orders of magnitudes. In practice, we used $\lambda_0=\mathcal{L}_{t0}/\mathcal{L}_{r0}$, where $\mathcal{L}_{t0}$ and $\mathcal{L}_{r0}$ were initial errors on the training set as per \eqref{eq:taskloss} and \eqref{eq:regularizationloss} by letting $c=0.5$. Following \cite{DeVries:2018}, we introduced a budget hyperparameter $\beta$ and allowed $\lambda$ to fluctuate within $(1\pm20\%)\lambda_0$ between iterations to prevent the unwished converges of $c$: if $\mathcal{L}_r>\beta$ then increased $\lambda$ (i.e., make it more expensive to ask for ``hints''), and if $\mathcal{L}_r<\beta$ then decreased $\lambda$ (i.e., make it more tolerant to produce small confidences). The budget $\beta$ was empirically set to 0.6. We observed that the selections of $\lambda_0$ and $\beta$ had no significant impact on the performance of final model.

In the validation phase, we uniformly sampled 12 square sub-images from a $4\times3$ grid on the full-resolution image and resized each of them to $224\times224$ (some full-resolution images would have unused margins if their aspect ratios are not $4:3$). For a naive network with illuminant estimation branch only, we inferred the global estimate by simply calculating the median over all local estimates. For a network with the confidence estimation branch, we used the normalized confidence scores as weights to aggregate local estimates:
\begin{equation}
\label{eq:aggregation}
\hat{\mathbf{L}}_{global} = \sum_i^{12}\frac{\hat{c}_i\cdot\hat{\mathbf{L}}_{local, i}}{\sum_i\hat{c}_i}\,.
\end{equation}

If an image produced confidence scores smaller than 0.5 for all the sub-images, naive aggregation method (calculating the median) would be used as the fallback.

Sub-images containing calibration targets (ColorChecker or SpyderCube) have been excluded for both training and validation.

We also present the analysis with respect to the space and time complexities for the proposed networks and some CNN-based models, as listed in Table~\ref{tab:complexity}. The results suggest that the proposed models are lightweight with lower computational and storage budgets compared to prior CNN-based models. When running the 3-Hierarchy model on a machine with Nvidia 1080-Ti GPU, the average inference time is 48ms/frame.

\begin{table}[!t]
\setlength\aboverulesep{0pt}
\setlength\belowrulesep{0pt}
\renewcommand{\arraystretch}{1.3}
\caption{The number of model parameters and operations (multiply-adds) in one forward propagation, assuming the input image has fixed size of 224$\times$224.}
\label{tab:complexity}
\centering
\begin{tabular}{p{.46\linewidth}|M{.16\linewidth}M{.17\linewidth}}
\toprule
Method & \#Param. & \#Ops.\\
\midrule
DS-Net$^{\text{a}}$~\cite{Shi:2016} & $\approx$17.3M & $\approx 6.0\times10^{10}$\\
Semantic CC~\cite{Mahmoud:2018} & $\approx$13.9M & $\approx 4.1\times10^9$\\
FC4 (AlexNet)~\cite{Hu:2017a} & $\approx$3.8M & $\approx 4.8\times10^9$\\
Deep Outdoor CC~\cite{Geoffroy:2017} & $\approx$3.7M & $\approx 2.3\times10^9$\\
Bianco CNN (2015)~\cite{Bianco:2015} & $\approx$ 154.4K & $\approx 7.3\times10^7$\\
Bianco CNN (2017)$^{\text{b}}$~\cite{Bianco:2017} & $\approx$ 154.7K & -\\
FFCC (Model M)$^{\text{c}}$~\cite{Barron:2017} & $\approx$ 24.6K & \cellcolor{2best}$\approx 3.4\times10^7$ \\
FFCC (Model Q)$^{\text{c}}$~\cite{Barron:2017} & $\approx$ 16.4K & $\cellcolor{3best}\approx 1.9\times10^6$ \\
CCC~\cite{Barron:2015} & \cellcolor{1best}$\approx$ 0.7K & \cellcolor{1best}$\approx 3.4\times10^7$\\
\hline
Ours, 1-Hierarchy, w/o conf. est.  & \cellcolor{2best}7.6K & $5.9\times10^7$\\
Ours, 1-Hierarchy, with conf. est.  & \cellcolor{3best}13.0K & $5.9\times10^7$\\
Ours, 2-Hierarchy, w/o conf. est.  & 32.7K & $1.9\times10^8$\\
Ours, 2-Hierarchy, with conf. est.  & 52.7K & $1.9\times10^8$\\
Ours, 3-Hierarchy, w/o conf. est.  & 112.6K & $4.7\times10^8$\\
Ours, 3-Hierarchy, with conf. est.  & 189.4K & $4.7\times10^8$\\
\bottomrule
\multicolumn{3}{p{.88\linewidth}}{\hspace{2ex}$^{\text{a}}$DS-Net accepts 44$\times$44 patches in the original paper. Enlarging the size of input will exponentially increase the amounts of operations for DS-Net. Nonetheless, to keep the comparison fair we fixed the sizes of inputs for all methods.}\\
\multicolumn{3}{p{.88\linewidth}}{\hspace{2ex}$^{\text{b}}$The number of operations for Bianco CNN (2017) is difficult to precisely calculate due to the complex processing pipelines.}\\
\multicolumn{3}{p{.88\linewidth}}{\hspace{2ex}$^{\text{c}}$The numbers of operations for FFCC variants only include chroma histograms construction and feature maps convolution, without taking into account the color space transform, bivariate von Mises fitting, Fourier regularization, and other post-processings.}
\end{tabular}
\end{table}

\section{Results}

\subsection{Intra-camera accuracy}\label{sec:intra-camera}

Intra-camera estimation accuracy was individually evaluated on 10 camera models from the ColorChecker RECommended and NUS-8 Camera datasets, and the results were reported by calculating the geometric means over all camera models from the same dataset. Tables~\ref{tab:ColorCheckerDataset} and \ref{tab:NUS-8} list the illuminant estimation accuracies of our models and prior mainstream algorithms. Following previous work, several standard metrics were reported in terms of angular error in degrees: mean, median, trimean, mean of the best quarter (best 25\%), and mean of the worst quarter (worst 25\%). It should be noted that according to~\cite{Hemrit:2018}, accuracies on the ColorChecker dataset presented in different papers are uncomparable bacause this dataset has at least 3 different sets of ground-truths. Therefore, we created Table~\ref{tab:ColorCheckerDataset} herein only for qualitative comparison purpose.

\begin{table}[!t]
\setlength\aboverulesep{0pt}
\setlength\belowrulesep{0pt}
\renewcommand{\arraystretch}{1.2}
\caption{Intra-camera illuminant color estimation accuracy (in degree) on the ColorChecker dataset~\cite{Hemrit:2018, Gehler:2016}.}
\label{tab:ColorCheckerDataset}
\centering
\begin{tabular}{p{.44\linewidth}|M{.05\linewidth}M{.05\linewidth}M{.05\linewidth}M{.05\linewidth}M{.05\linewidth}}
\toprule
Method & Mean & Med. & Tri. & Best 25\% & Worst 25\%\\
\midrule
White-Patch~\cite{Brainard:1986} & 7.55 & 5.68 & 6.35 & 1.45 & 16.12\\
Gray-World~\cite{Buchsbaum:1980} & 6.36 & 6.28 & 6.28 & 2.33 & 10.58\\
1st-order Gray-Edge~\cite{Weijer:2007a} & 5.33 & 4.52 & 4.73 & 1.86 & 10.03\\
Shades-of-Gray~\cite{Finlayson:2004} & 4.93 & 4.01 & 4.23 & 1.14 & 10.20\\
Bayesian~\cite{Gehler:2008} & 4.82 & 3.46 & 3.88 & 1.26 & 10.49\\
Spatio-spectral Statistics~\cite{Chakrabarti:2011} & 3.59 & 2.96 & 3.10 & 0.95 & 7.61\\
Pixels-based Gamut~\cite{Barnard:2000} & 4.20 & 2.33 & 2.91 & 0.50 & 10.72\\
DGP~\cite{Qian:2018} & 3.07 & 1.87 & 2.16 & 0.43 & 7.62\\
Quasi-Unsupervised (in-db)~\cite{Bianco:2019} & 2.91 & 1.98 & - & - & -\\
Bianco CNN (2015)~\cite{Bianco:2015} & 2.63 & 1.98 & 2.10 & 0.72 & \cellcolor{2best}3.90\\
Cheng et al. 2015~\cite{Cheng:2015} & 2.42 & 1.65 & 1.75 & 0.38 & 5.87\\
FFCC (Model Q)~\cite{Barron:2017} & 2.01 & 1.13 & 1.38 & 0.30 & 5.14\\
CCC~\cite{Barron:2015} & 1.95 & 1.22 & 1.38 & \cellcolor{3best}0.35 & 4.76\\
DS-Net~\cite{Shi:2016} & 1.90 & \cellcolor{2best}1.12 & \cellcolor{3best}1.33 & \cellcolor{2best}0.31 & 4.84\\
FFCC (Model M)~\cite{Barron:2017} & \cellcolor{2best}1.78 & \cellcolor{1best}0.96 & \cellcolor{1best}1.14 & \cellcolor{1best}0.29 & 4.62\\
FC4 (SqueezeNet)~\cite{Hu:2017a} & \cellcolor{1best}1.65 & \cellcolor{3best}1.18 & \cellcolor{2best}1.27 & 0.38 & \cellcolor{1best}3.78\\
\hline
Ours, 1-Hierarchy, w/o conf. est. & 2.41 & 2.02 & 2.00 & 0.59 & 5.10\\
Ours, 1-Hierarchy, with conf. est. & 2.32 & 1.96 & 1.96 & 0.60 & 4.65\\
Ours, 2-Hierarchy, w/o conf. est. & 2.18 & 1.73 & 1.82 & 0.53 & 4.70\\
Ours, 2-Hierarchy, with conf. est. & 2.10 & 1.68 & 1.77 & 0.49 & 4.32\\
Ours, 3-Hierarchy, w/o conf. est. & 1.98 & 1.38 & 1.52 & 0.51 & 4.52\\
Ours, 3-Hierarchy, with conf. est. & \cellcolor{3best}1.85 & 1.31 & 1.37 & 0.44 & \cellcolor{3best}4.14\\
\bottomrule
\end{tabular}
\end{table}

\begin{table}[!t]
\setlength\aboverulesep{0pt}
\setlength\belowrulesep{0pt}
\renewcommand{\arraystretch}{1.2}
\caption{Intra-camera illuminant color estimation accuracy (in degree) on the NUS-8 Camera dataset~\cite{Cheng:2014}.}
\label{tab:NUS-8}
\centering
\begin{tabular}{p{.44\linewidth}|M{.05\linewidth}M{.05\linewidth}M{.05\linewidth}M{.05\linewidth}M{.05\linewidth}}
\toprule
Method & Mean & Med. & Tri. & Best 25\% & Worst 25\%\\
\midrule
White-Patch~\cite{Brainard:1986} & 10.62 & 10.58 & 10.49 & 1.86 & 19.45\\
Pixel-based Gamut~\cite{Barnard:2000} & 7.70 & 6.71 & 6.90 & 2.51 & 14.05\\
Gray-World~\cite{Buchsbaum:1980} & 4.14 & 3.20 & 3.39 & 0.90 & 9.00\\
Bayesian~\cite{Gehler:2008} & 3.67 & 2.73 & 2.91 & 0.82 & 8.21\\
Shades-of-Gray~\cite{Finlayson:2004} & 3.40 & 2.57 & 2.73 & 0.77 & 7.41\\
1st-order Gray-Edge~\cite{Weijer:2007a} & 3.20 & 2.22 & 2.43 & 0.72 & 7.36\\
Spatio-spectral Statistics~\cite{Chakrabarti:2011} & 2.96 & 2.33 & 2.47 & 0.80 & 6.18\\
Cheng et al. 2014~\cite{Cheng:2014} & 2.92 & 2.04 & 2.24 & 0.62 & 6.61\\
DGP~\cite{Qian:2018} & 2.91 & 1.97 & 2.13 & 0.56 & 6.67\\
DS-Net~\cite{Shi:2016} & 2.24 & \cellcolor{3best}1.46 & 1.68 & 0.48 & 6.08\\
CCC~\cite{Barron:2015} & 2.38 & 1.48 & 1.69 & \cellcolor{3best}0.45 & 5.85\\
SqueezeNet-FC4~\cite{Hu:2017a} & 2.23 & 1.57 & 1.72 & 0.47 & 5.15\\
Cheng et al. 2015~\cite{Cheng:2015} & 2.18 & 1.48 & 1.64 & 0.46 & \cellcolor{2best}5.03\\
FFCC (Model M)~\cite{Barron:2017} & \cellcolor{2best}1.99 & \cellcolor{1best}1.31 & \cellcolor{1best}1.43 & \cellcolor{1best}0.35 & \cellcolor{1best}4.75\\
Quasi-Unsupervised (in-db)~\cite{Bianco:2019} & \cellcolor{1best}1.97 & \cellcolor{2best}1.41 & - & - & -\\
\hline
Ours, 1-Hierarchy, w/o conf. est. & 2.84 & 1.92 & 2.04 & 0.80 & 5.82\\
Ours, 1-Hierarchy, with conf. est. & 2.84 & 1.88 & 1.90 & 0.75 & 5.39 \\
Ours, 2-Hierarchy, w/o conf. est. & 2.32 & 1.64 & 1.67 & 0.46 & 5.44\\
Ours, 2-Hierarchy, with conf. est. & 2.27 & 1.61 & \cellcolor{3best}1.63 & 0.48 & 5.16\\
Ours, 3-Hierarchy, w/o conf. est. & \cellcolor{3best}2.18 & 1.59 & 1.74 & 0.48 & 5.35\\
Ours, 3-Hierarchy, with conf. est. & 2.20 & 1.53 & \cellcolor{2best}1.60 & \cellcolor{2best}0.44 & \cellcolor{3best}5.07\\
\bottomrule
\end{tabular}
\end{table}

It is worth mentioning that with the confidence estimation branch, noticeable improvements are observed for the worst 25\% metrics on both datasets, which indicates that the confidence estimation branch provides useful information for the local estimates aggregation and makes the model more robust to the estimation ambiguities in hard cases.

\subsection{Inter-camera accuracy}

We evaluated inter-camera estimation accuracy on the MultiCam dataset using leave-one-out cross validation. Table~\ref{tab:MultiCam} lists the comparison of the proposed models with other algorithms, in which all the metrics for the learning-based (camera-known) algorithms were obtained by calculating the arithmetic means over all rounds of the leave-one-out cross validations (with asterisk superscripts) as discussed in subsection~\ref{sec:benchmark_datasets}, and all the metrics for the assumption-based (camera-agnostic) ones were obtained by directly analyzing over all images from the MultiCam dataset (without asterisks).

\begin{table}[!t]
\setlength\aboverulesep{0pt}
\setlength\belowrulesep{0pt}
\renewcommand{\arraystretch}{1.2}
\caption{Inter-camera illuminant color estimation accuracy (in degree) on the MultiCam dataset, computed in the individual camera color spaces. For the camera-known algorithms (with asterisks), the results were obtained by calculating the arithmetic means over all rounds of the cross validation; for the camera-agnostic methods (without superscripts), the results were obtained by directly taking the average from all images in the MultiCam dataset.}
\label{tab:MultiCam}
\centering
\begin{tabular}{p{.44\linewidth}|M{.05\linewidth}M{.05\linewidth}M{.05\linewidth}M{.05\linewidth}M{.05\linewidth}}
\toprule
Method & Mean & Med. & Tri. & Best 25\% & Worst 25\%\\
\midrule
Gray-World~\cite{Buchsbaum:1980} & 4.57 & 3.63 & 3.85 & 1.04 & 9.64\\
Pixel-based Gamut$^\ast$~\cite{Barnard:2000} & 3.76 & 2.99 & 3.10 & 1.14 & 7.70\\
White-Patch~\cite{Brainard:1986} & 3.64 & 2.84 & 2.95 & 1.17 & 7.48\\
1st-order Gray-Edge~\cite{Weijer:2007a} & 3.21 & 2.51 & 2.65 & 0.93 & 6.61\\
2st-order Gray-Edge~\cite{Weijer:2007a} & 3.12 & 2.42 & 2.54 & 0.86 & 6.55\\
Bayesian$^\ast$~\cite{Gehler:2008} & 3.04 & 2.28 & 2.40 & 0.67 & 6.69\\
Shades-of-Gray~\cite{Finlayson:2004} & 2.93 & 2.24 & 2.41 & 0.66 & 6.31\\
Spatio-spectral Statistics$^\ast$~\cite{Chakrabarti:2011} &  2.92 & 2.08 & 2.17 & 0.46 & 6.50\\
DGP~\cite{Qian:2018} & 2.80 & 2.00 & 2.22 & 0.55 & 6.25\\
Quasi-Unsupervised (no-db)~\cite{Bianco:2019} & 2.39 & 1.69 & 1.89 & 0.48 & 5.47\\
Bianco CNN (2015)$^\ast$~\cite{Bianco:2015} & 1.88 & 1.47 & 1.54 & 0.38 & 4.90\\
FFCC (Model M)$^\ast$~\cite{Barron:2017} & \cellcolor{2best}1.55 & 1.22 & \cellcolor{2best}1.23 & \cellcolor{1best}0.32 & \cellcolor{3best}3.66\\
SqueezeNet-FC4$^\ast$~\cite{Hu:2017a} & \cellcolor{1best}1.54 & \cellcolor{1best}1.13 & \cellcolor{1best}1.20 & \cellcolor{2best}0.32 & \cellcolor{2best}3.59\\
\hline
Ours, 1-Hierarchy, w/o conf. est.$^\ast$  & 2.03 & 1.51 & 1.64 & 0.43 & 4.47\\
Ours, 1-Hierarchy, with conf. est.$^\ast$ & 2.07 & 1.54 & 1.64 & 0.43 & 4.41\\
Ours, 2-Hierarchy, w/o conf. est.$^\ast$  & 1.78 & 1.40 & 1.47 & 0.40 & 4.05\\
Ours, 2-Hierarchy, with conf. est.$^\ast$ & 1.77 & 1.40 & 1.46 & 0.38 & 3.83\\
Ours, 3-Hierarchy, w/o conf. est.$^\ast$  & 1.67 & \cellcolor{2best}1.20 & \cellcolor{3best}1.30 & \cellcolor{3best}0.38 & 3.78\\
Ours, 3-Hierarchy, with conf. est.$^\ast$ & \cellcolor{3best}1.64 & \cellcolor{3best}1.22 & 1.30 & 0.40 & \cellcolor{1best}3.54\\
\bottomrule
\end{tabular}
\end{table}

It can be observed in Table~\ref{tab:MultiCam} that the inter-camera estimation accuracies are even slightly better than those in the intra-camera evaluations. We reckon that this improvement is mainly due to two reasons: a) the inclusion of Cube dataset lowers the average errors of the cross validation because images in this dataset are relatively "unchallenged" compared to ColorChecker RECommended and NUS-8 Camera datasets, which was also indicated in~\cite{Banic:2017}, and b) the increasing number of training images from the MultiCam dataset enhances the models' learning capacities and help them avoid overfitting, especially for those deep-learning-based models.

It should also be noted that the inter-camera experiments we performed were neither technically \textit{camera-agnostic} nor \textit{unsupervised} since the color correction matrices were device-dependent and had been manually selected, and thus the results in Table~\ref{tab:MultiCam} shall not be comparable to those reported in the prior works focusing on the unsupervised learning or the camera-agnostic scenario~\cite{Bianco:2019, Qian:2019, Qian:2018}.

\subsection{Visualization}

\begin{figure*}[!h]
\centering
\includegraphics[width=\linewidth]{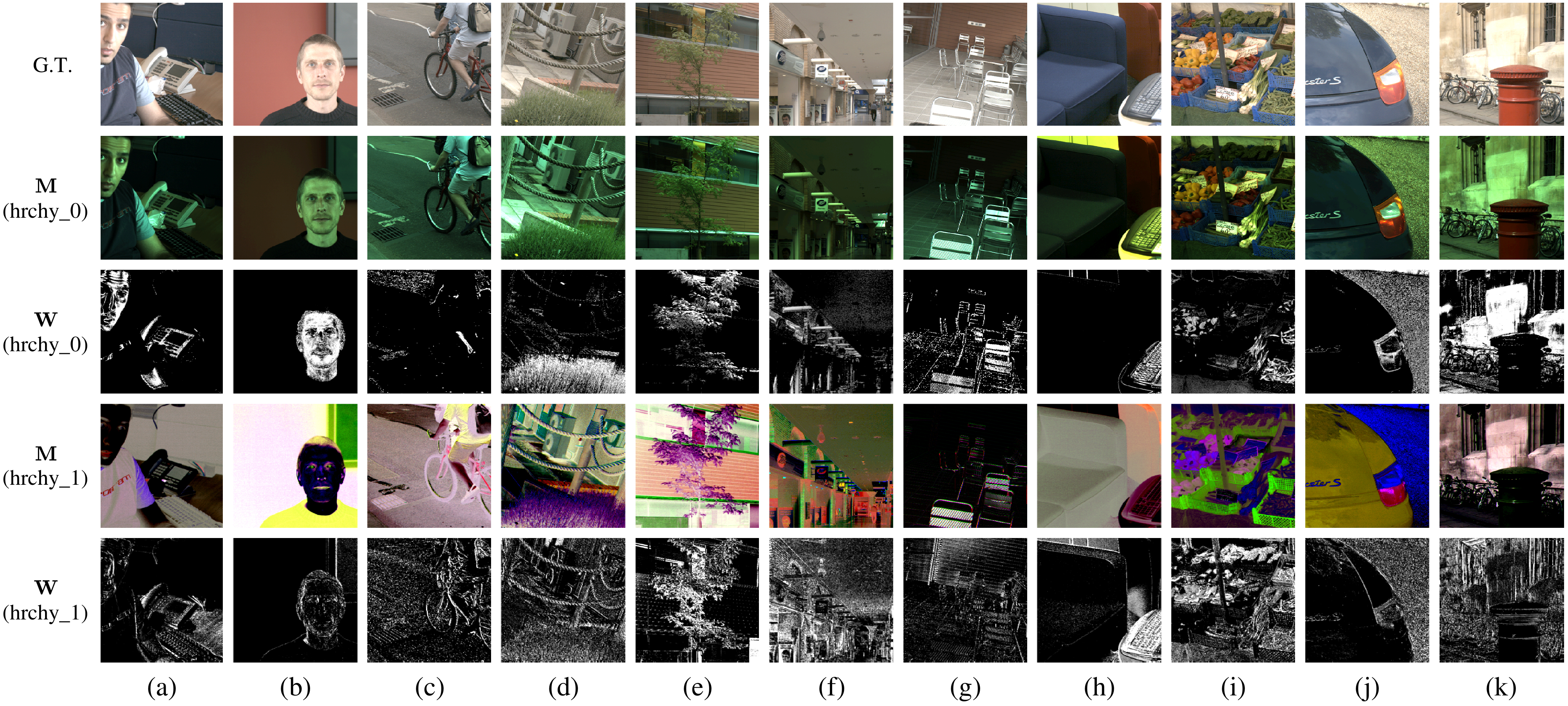}
\caption{Examples of input feature maps and reweighting maps of ReWUs in the naive 1-Hierarchy network, trained on the ColorChecker RECommended dataset~\cite{Hemrit:2018}. \textbf{From top to bottom:} the ground truth images (without color correction), the input color biased images (also the input feature maps of the ReWU in ``hrchy\_0''), the reweighting maps of the ReWU in ``hrchy\_0'', the input feature maps of the ReWU in ``hrchy\_1'', and the reweighting maps of the ReWU in ``hrchy\_1''. All feature maps have been normalized for visualization. The input feature maps $\mathbf{M}$ of ``hrchy\_1'' (fourth row) have 16 channels, here only 3 channels out of 16 with maximum activations are shown.}
\label{fig:visualization}
\end{figure*}

Figure~\ref{fig:visualization} presents some examples to visualize how the ReWUs work and what it has learned. As expected, the ReWU connected to the input images highlights achromatic pixels with moderate brightness (subfigures a and f), the surfaces with high reflectance (subfigures g and h), as well as some memory colors with high frequency of occurrences like skin tone (subfigures a--c) and green on the plants (subfigures d, e, and i). In the other hand, ReWU also suppresses those eccentric colors that may lead to the prejudicial decisions for the illuminant estimation, for example, the red paint on the wall (subfigure b), the orange on the peppers (subfigure i), the navy blue on the roadster (subfigure j), and the scarlet on the pillar box (subfigure k). The great interpretability of the ReWU makes it possible to serve as a guidance for designing statistics-based methods, especially for those low-end devices that do not have sufficient computational capacity to run CNN-based models.

\begin{figure}[!t]
\centering
\includegraphics[width=.75\linewidth]{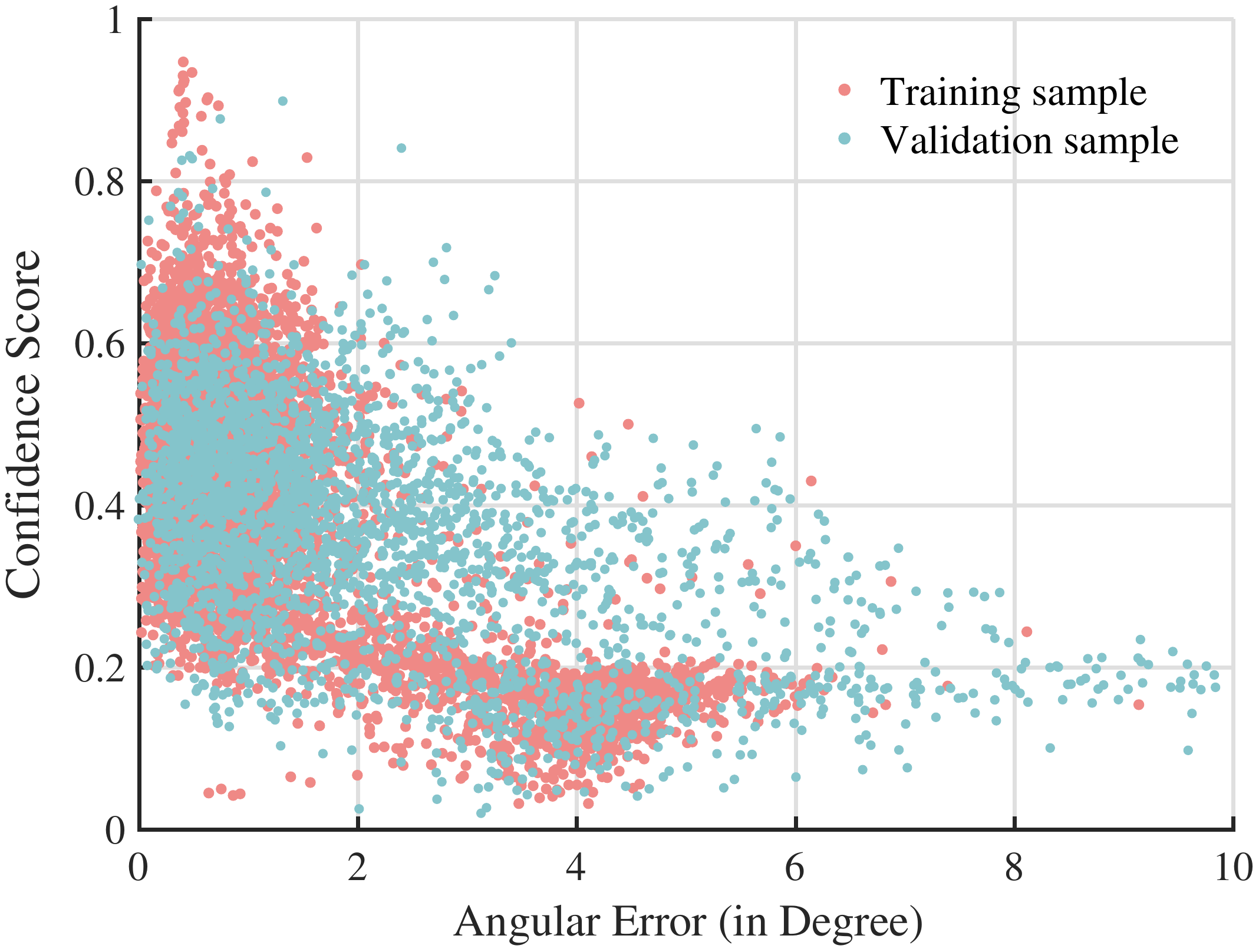}
\caption{The confidence scores produced by the confidence estimation branch versus angular errors of the illuminant color estimation. 3-Hierarchy architecture was used in this experiment, with $\lambda_0=8\times10^{-5}$ and $\beta=0.6$. The model was trained on the MultiCam dataset, which contains 57,974 samples for training and 28,985 for validation after data augmentation and sub-image cropping. For clarity, only 10\% samples were randomly picked up and plotted.}
\label{fig:angular_err_vs_confidence}
\end{figure}

To investigate how the confidence estimation branch helps to improve the reliability of local estimates aggregation, we draw a scatterplot of the angular errors of local estimates with respect to the confidence scores for all sub-images in the MultiCam dataset, as shown in \figurename~\ref{fig:angular_err_vs_confidence}. As expected, the confidence estimation branch has successfully learned to assign high confidence scores for images that can be accurately estimated, and low scores for those that are prone to be erroneous. It can also be observed in \figurename~\ref{fig:angular_err_vs_confidence} that some images with very low angular error (even $0^\circ$) are assigned with low confidence scores, which we assume is due to the loss function and the training strategy we used---the confidence estimation branch is prone to penalize those incorrectly predicted images by assigning them with low confidence socres, whilst be more ``tolerant'' to cases where unconfident images are actually well-predicted.

\begin{figure}[!t]
\centering
\subfloat{\includegraphics[width=.9\linewidth]{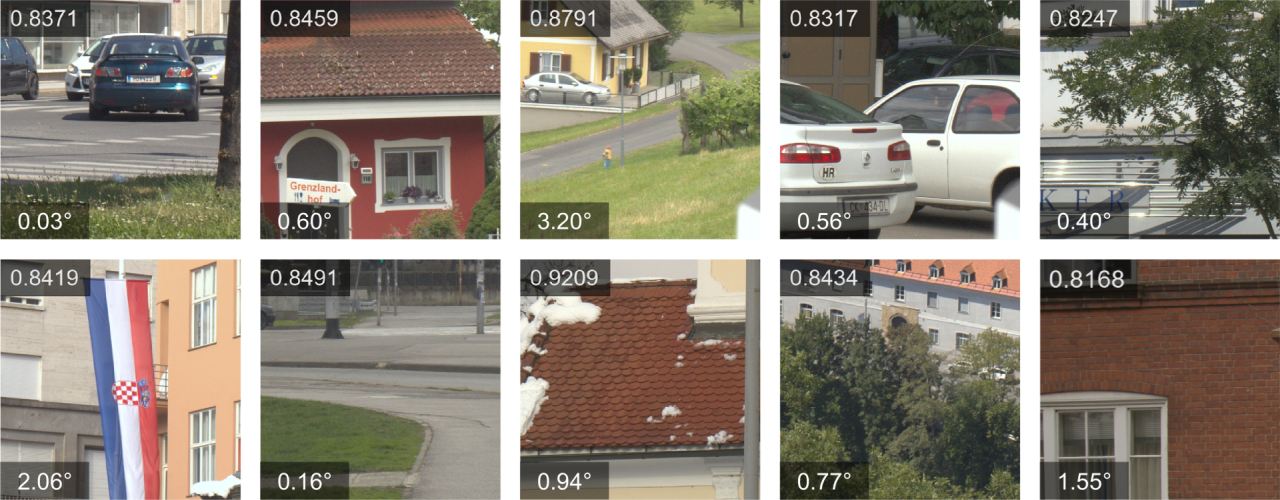}}\\[1em]
\hrule
\subfloat{\includegraphics[width=.9\linewidth]{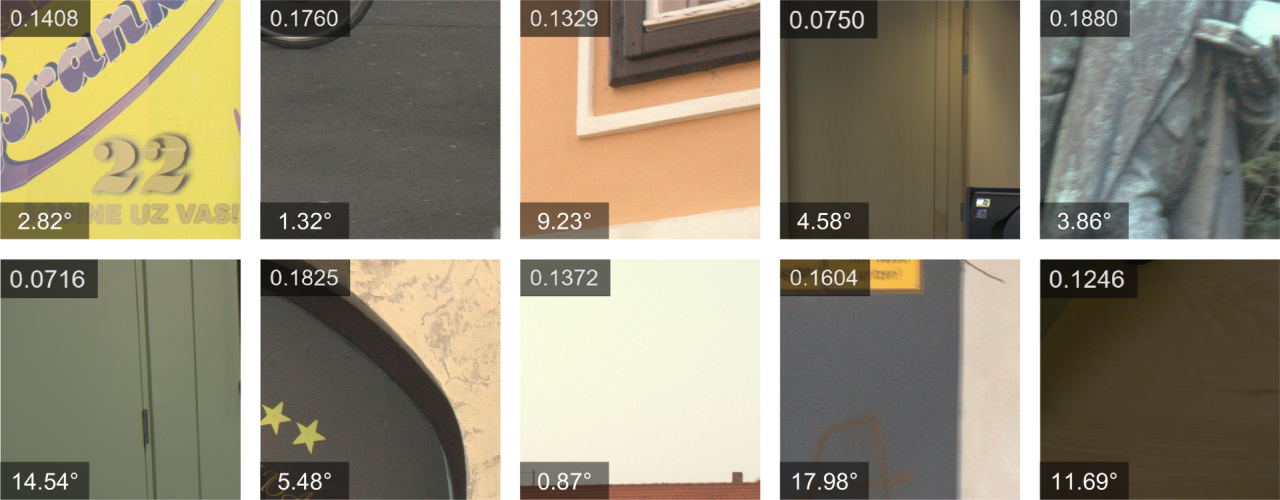}}
\caption{Typical sub-images in the MultiCam dataset with high (above the separator) and low (beneath the separator) predicted confidence scores. For those challenging sub-images with ambiguous patterns, although the estimation accuracies are quite poor (see the relatively large augular errors in the left-bottom corner of sub-images), the lower confidence scores (in the left-top corner of sub-images) help to reduce the risks aggregating them into the global estimate.}
\label{fig:confidences}
\end{figure}

Furthermore, we sorted all the sub-images in the validation set by their confidence scores and surprisedly found that decision-making behavior of the confidence estimation branch was very similar to human's color constancy mechanisms: when the scenes contained valuable clues for estimating illuminant color, e.g., objects with memory colors and/or recognizable patterns, the network became more confident about its estimates; oppositely, when the objects and/or colors in the scenes were difficult to identify, the network appeared ambiguous about its decisions and thus produced low confidence scores. Figure~\ref{fig:confidences} demonstrates some typical sub-images from the validation set with different levels of uncertainties, in which scenes containing frequently-occurring objects (lawns, neutral surfaces, brick roofs, etc.) are categorized as ``strongly confident to estimate'', and scenes with vague patterns (pseudo-neutral objects, surfaces with a dominant color, etc.) are categorized as ``unconfident to estimate''.

\section{Conclusion}

A novel illuminant color estimation framework is proposed to exploit useful cues from feature maps in an efficient and interpretable way. To quantify the confidences of local illuminant estimates, an uncertainty prediction branch is included, upon which more flexible decisions can be made to determine the illuminant color. Experimental results indicate that the proposed method achieves comparative performance with other state-of-the-art high-level models for most of metrics with a more compact model size and lower computational burden, making it suitable for applications deployed on mobile platforms. In addition to being employed for computational color constancy, the image feature reweight unit (ReWU) also exhibits reasonable potential for other color-relevant applications such as fine-grained classification and image semantic segmentation. As the future work we plan to investigate if the confidence estimation branch could be improved so as to reduce the occurrence of ``false negative'' samples, i.e. images with good predictions but assigned with low confidence score, possibly by using some reinforcement learning strategy or by proposing a better loss function.

\appendix[Proof of Equation (17)]
\begin{IEEEproof}
For the sake of briefness, let $X$ and $Y$ be the random variables of values in the channel axes of $C\!N(\mathbf{g}\ast\mathbf{M})$ and $min(C\!N(\mathbf{g}\ast\mathbf{M}))$, respectively.

By definition, the distribution function for $Y$ is calculated as
\begin{equation}
\label{eq:distributionfunc1}
\begin{aligned}
F(y)&=P(Y\leq{}y)=1-P(Y>y)\\
    &=1-P\left(min(X_1,X_2,\ldots,X_{K})>y\right)\,,
\end{aligned}
\end{equation}
where the subscripts are channel indices, $K$ is the number of channels in $C\!N(\mathbf{g}\ast\mathbf{M})$. Since all values in $X$ are i.i.d., we have
\begin{equation}
\label{eq:distributionfunc2}
\begin{aligned}
F(y)&=1-P(X_1>y)P(X_2>y)\ldots{}P(X_{K}>y)\\
	&=1-P(X_1>y)^{K}\,.
\end{aligned}
\end{equation}

According to the definition of the channel normalization in~\eqref{eq:channelnormalization}, $X$ approximately follows the standard normal distribution, which yields
\begin{equation}
\label{eq:distributionfunc3}
P(X_1>y) = 1-P(X_1\leq{}y) = 1-\Phi(y;0,1)\,,
\end{equation}
where $\Phi(y;0,1)$ is the cumulative distribution function of the standard normal distribution. Therefore,
\begin{equation}
\label{eq:distributionfunc2}
F(y)=1-P(X_1>y)^{K}=1-\big(1-\Phi(y;0,1)\big)^{K}\,.
\end{equation}

The expectation of $min(C\!N(\mathbf{g}\ast\mathbf{M}))$ can be calculated as
\begin{equation}
\label{eq:integration}
\begin{aligned}
&\E[min(C\!N(\mathbf{g}\ast\mathbf{M}))]=\E[Y]\\[.5em]
=&\int_{-\infty}^\infty{}yf(y)\D{y}=\int_{-\infty}^\infty{}yF^\prime(y)\D{y}\\[.2em]
=&K\int_{-\infty}^\infty{}y\,\varphi(y;0,1)\big(1-\Phi(y;0,1)\big)^{K-1}\D{y}\,,
\end{aligned}
\end{equation}
where $\varphi(y;0,1)$ is the probability density function of the standard normal distribution.
\end{IEEEproof}

% Can use something like this to put references on a page
% by themselves when using endfloat and the captionsoff option.
\ifCLASSOPTIONcaptionsoff
  \newpage
\fi

\section*{Acknowledgment}

This research was supported by Zhejiang Province Foundation for Cultural Heritage Preservation Technology (2017010) and Fundamental Research Funds for the Central Universities (2018FZA128).

% trigger a \newpage just before the given reference
% number - used to balance the columns on the last page
% adjust value as needed - may need to be readjusted if
% the document is modified later
%\IEEEtriggeratref{8}
% The "triggered" command can be changed if desired:
%\IEEEtriggercmd{\enlargethispage{-5in}}

% references section

\bibliographystyle{IEEEtran}
\bibliography{IEEEabrv,reweight_cc}

% biography section
% 
% If you have an EPS/PDF photo (graphicx package needed) extra braces are
% needed around the contents of the optional argument to biography to prevent
% the LaTeX parser from getting confused when it sees the complicated
% \includegraphics command within an optional argument. (You could create
% your own custom macro containing the \includegraphics command to make things
% simpler here.)
%\begin{IEEEbiography}[{\includegraphics[width=1in,height=1.25in,clip,keepaspectratio]{mshell}}]{Michael Shell}
% or if you just want to reserve a space for a photo:

\begin{IEEEbiography}[{\includegraphics[height=1.25in,clip,keepaspectratio]{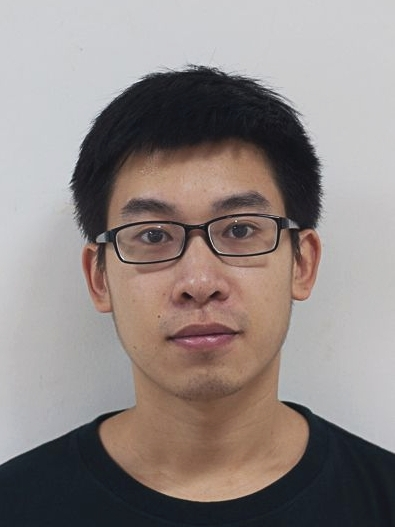}}]{Jueqin Qiu}
received his B.S. degree in optical engineering from Beijing University of Aeronautics and Astronautics, China, in 2014. He is currently pursuing the Ph.D. degree in College of Optical Science and Engineering at Zhejiang University. His current research interests include computer vision, digital imaging processing, and color science.
\end{IEEEbiography}

\begin{IEEEbiography}[{\includegraphics[height=1.25in,clip,keepaspectratio]{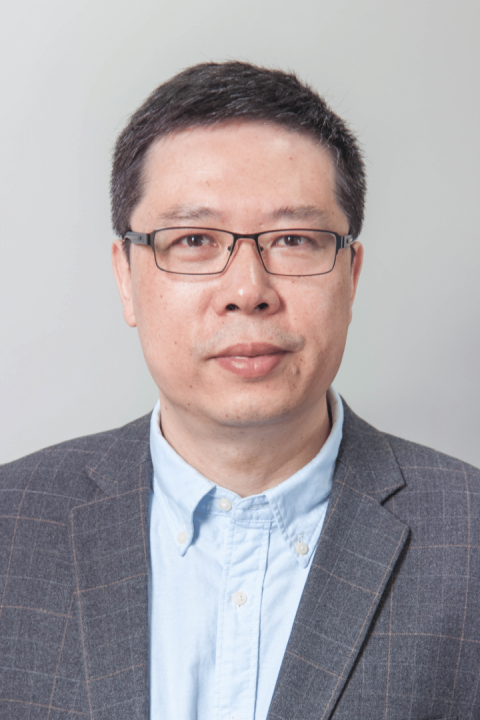}}]{Haisong Xu}
received his Ph.D. degree in optical engineering from Zhejiang University, China, in 1993. He was a postdoctoral research fellow in Zhejiang University from 1994 to 1995 and in Chiba University, Japan, from 1999 to 2001. Now he is a professor in College of Optical Science and Engineering at Zhejiang University. His current research interests include color science, imaging technology, and lighting engineering. He is the China representative in D1 of CIE, in AIC, and in ACA.
\end{IEEEbiography}

\begin{IEEEbiography}[{\includegraphics[height=1.25in,clip,keepaspectratio]{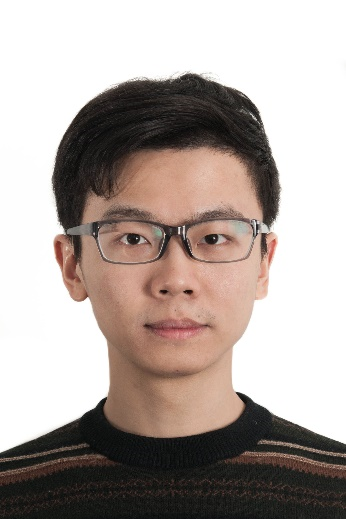}}]{Zhengnan Ye}
received his B.S. degree in optical engineering from Zhejiang University, China, in 2015. He is now studying for the Ph.D. degree in College of Optical Science and Engineering at Zhejiang University. His research interests includes color science and multispectral imaging technology.
\end{IEEEbiography}

% You can push biographies down or up by placing
% a \vfill before or after them. The appropriate
% use of \vfill depends on what kind of text is
% on the last page and whether or not the columns
% are being equalized.

\vfill

% Can be used to pull up biographies so that the bottom of the last one
% is flush with the other column.
%\enlargethispage{-5in}

% that's all folks
\end{document}